\documentclass[11pt,a4paper]{article}
\usepackage[hyperref]{paper}
\usepackage{times}
\usepackage{latexsym}
\usepackage{graphicx}
\usepackage{multirow,color} 
\usepackage{helvet}  
\usepackage{courier}  
\usepackage{url} 
\usepackage{epstopdf}

\aclfinalcopy 

\title{Implicit Regularization of Stochastic Gradient Descent in Natural Language Processing: Observations and Implications}

\author{Deren Lei, Zichen Sun, Yijun Xiao, William Yang Wang\\
Department of Computer Science\\
University of California, Santa Barbara\\
\{derenlei, zichen\_sun\}@ucsb.edu, \{yijunxiao, william\}@cs.ucsb.edu\\
}

\date{}
\begin{document}
\maketitle
\begin{abstract}
Deep neural networks with remarkably strong generalization performances are usually over-parameterized. Despite explicit regularization strategies are used for practitioners to avoid over-fitting, the impacts are often small. Some theoretical studies have analyzed the implicit regularization effect of stochastic gradient descent (SGD) on simple machine learning models with certain assumptions. However, how it behaves practically in state-of-the-art models and real-world datasets is still unknown. To bridge this gap, we study the role of SGD implicit regularization in deep learning systems. We show pure SGD tends to converge to minimas that have better generalization performances in multiple natural language processing (NLP) tasks. This phenomenon coexists with dropout, an explicit regularizer. In addition, neural network's finite learning capability does not impact the intrinsic nature of SGD's implicit regularization effect. Specifically, under limited training samples or with certain corrupted labels, the implicit regularization effect remains strong. We further analyze the stability by varying the weight initialization range. We corroborate these experimental findings with a decision boundary visualization using a 3-layer neural network for interpretation. Altogether, our work enables a deepened understanding on how implicit regularization affects the deep learning model and sheds light on the future study of the over-parameterized model's generalization ability.
\end{abstract}

\noindent 

\section{Introduction}
Modern machine learning systems based on deep neural networks, which are usually over-parameterized\footnote{e.g., Those models that often have far more parameters than the number of training examples.} \cite{soltanolkotabi2018theoretical,ari2009search,denton2014exploiting} , are prone to over-fitting. In practice, there exists several strategies that explicitly help models achieve better generalization including \(L_{1}\), \(L_{2}\) regularization of parameters. There also exists algorithmic ways to achieve the same goal such as dropout \cite{srivastava2014dropout}. Recent studies show that these explicit regularization techniques do not significantly affect the generalization performance, but the model architecture itself takes a more important role \cite{zhang2016understanding}. Several theory papers discover that the SGD has an implicit regularization effect that tends to make the norm of weight smaller on logistic regression and 1-layer neural network \cite{li2017algorithmic}. Understanding this effect in practical deep learning systems is crucial.

\subsection{Our Contributions}
In this paper, we study how implicit regularization behaves in deep learning systems for NLP tasks and how the implications are related to understanding over-parameterized model's generalization performance. Specifically, we try to answer the following research questions:

\smallskip
\noindent \textit{\textbf{RQ1:} Does SGD implicit regularization exists in deep learning systems? If so, how can we make better use of this phenomenon in the state-of-the-art models?}
\smallskip

\subsubsection*{SGD and implicit regularization} 
We show that the implicit regularization effect of SGD will gradually disappear as pure SGD move towards the mini-batch methods. Since pure SGD converges slowly, we recommend researchers with abundant computational resource use pure SGD instead of mini-batch to achieve a potentially better performance.
\subsubsection*{The role of explicit regularization} 
We discover that the implicit regularization effect is observable even adding the dropout, although such explicit regularizer can compensate some generalization gaps for large batch sizes. It gives us a better understanding of why explicit regularization does not help that much.

\smallskip
\noindent \textit{\textbf{RQ2:} Will SGD implicit regularization be affected by model's capability of finding the intrinsic patterns, especially when training data is challenging to learn?}
\smallskip

\subsubsection*{Impact of limited training data}
Neural networks have their own limited capability to learn based on the training data size. Intuition suggests that if the dataset size is extremely small, the neural networks will struggle to learn the intrinsic pattern of the data, e.g. extremely bad generalization performance. To our surprise, we only observe a slow steady decrease of the performance as we decrease the dataset size to \(10\%\) of the original. More than that, we observe the implicit regularization effect under different kind of partial dataset sizes. It means that the implicit regularization as the intrinsic nature of SGD will only be affected by the dataset sizes for a small amount if the model is over-parameterized. It sheds light on future analysis upon the generalization performance of over-parameterized models.

\subsubsection*{Fitting corrupted labels}
Over-parameterized models are able to learn random labels: they tend to learn the recognizable patterns while at the same time fit noises by brute-force \cite{zhang2016understanding}. In this case, the pattern is becoming harder to recognize and the converging of SGD is slower while the noise level becomes higher. Under this situation, is the decrease of generalization performance due to the decline of implicit regularization effect? Our experiments surprisingly show that under different acceptable noise levels, the implicit regularization effects are the same. It helps researchers study why over-parameterized model have generalization ability even though it can fit random data better. 

\smallskip
\noindent \textit{\textbf{RQ3:} Initialization of the model will affect its performance. Can we discover any correlation between initialization range and implicit regularization performance?}
\smallskip

\subsubsection*{Initialization based effect}
Recent studies show that smaller weight initialization provides better generalization on the over-parameterized matrix factorization models using gradient descent (GD) with early stopping \cite{li2017algorithmic}. Considering about initialization might potentially impact the implicit regularization effect when doing experiments, yet, we discover that this conclusion is not universal in deep learning systems with SGD. Our experiments expose that models' implicit regularization performances do not have a steady monotonous relationship with the weight initialization range. Base on our results, higher implicit regularization performance models tend to prefer smaller initialization ranges. We will illustrate our findings in the following sections.

\section{NLP Tasks and Configuration}
\begin{figure*}[t]
\centering
\includegraphics[width=15.5cm]{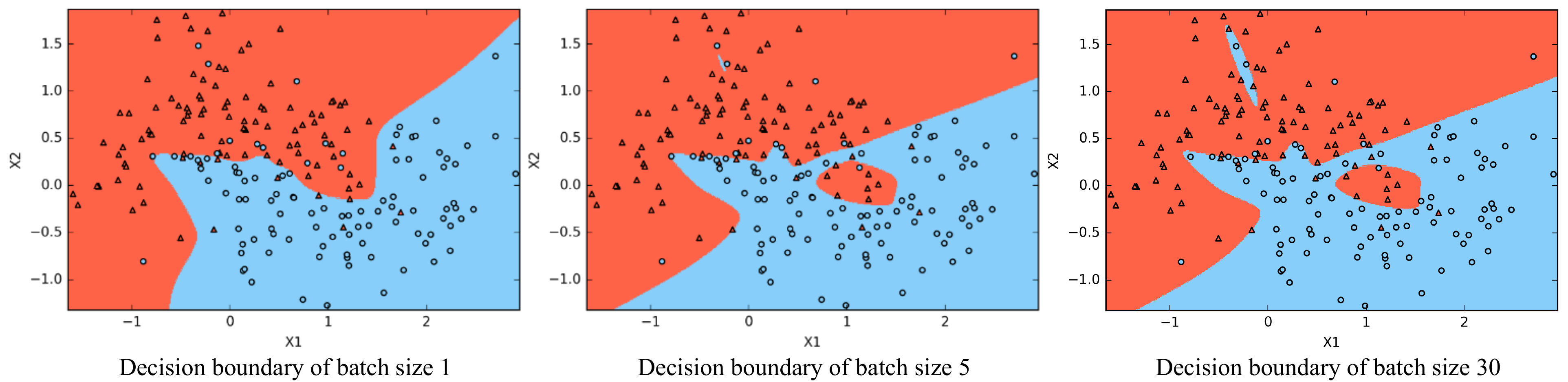}
\caption{The decision boundary of shallow neural network model on a synthetic dataset with a train/test split ratio 7:3. Larger batch sizes tend to be more complex while all achieve a test accuracy 91.67.}
\label{fig:decision}
\end{figure*}
In order to observe the impact of SGD implicit regularization, we use the vanilla SGD with batch size 1 as the baseline method. Then we observe how the model performs as we increase the batch size to shift to GD. Since different settings should be treated equally, instead of iterating through training data by epochs, we shuffle the data each time a random batch is selected, and measure the number of steps required. We choose sentence classification, language modeling, and sequence tagging as our target NLP tasks since they are core NLP problems with different machine learning settings. Our reported results are averaged over at least 5 runs. We specifically focus on SGD without momentum. Given a random pair \((x^{(i)}, y^{(i)})\) from the training data, the new update of a parameter \(\theta\) with learning rate \(\eta\) is
\begin{equation}
\theta := \theta - \eta \nabla_{\theta} Q(\theta; x^{(i)}, y^{(i)})
\end{equation}
where \(Q(\theta)\) is the loss function and \(\hat{y}\) stands for the computed label. Our experiments specifically focus on cross-entropy loss because it is practical but lack of theory construction in SGD implicit regularization:
\begin{equation}
Q(\theta) = -y^{(i)}ln(\hat{y}^{(i)})
\end{equation}
Note that the definition above is only for pure SGD that update each parameter by the batch size of 1. Consider that larger batch size can also be used for the optimization method, the update rule with batch size \(n\) can be generalized to
\begin{equation}
\theta := \theta - \frac{\eta}{n} \sum_{i=1}^{n}  \nabla_{\theta} Q(\theta; x^{(i)}, y^{(i)})
\end{equation}
As the batch size increases from 1 to the size of the complete training sample, the optimization method will gradually move from pure SGD to GD. If the implicit regularization of pure SGD can be observed, the generalization performance will decrease as batch size \(n\) in the above update rule increases. We define the degrade of generalization performance as: the performance decreases while batch size increases. This technique is used to study SGD implicit regularization.

\subsection{Sentence classification}
The first task that we investigate is sentence classification. The goal is to classify the sentiment of a sentence as positive or negative with a one-hot label \([y_{1}, y_{2}]\), where \(y_{1}\) stands for probability positive meaning and \(y_{2}\) stands for the contrary. The model will learn the probability given a sentence \(x = (x^{1}, x^{2},...,x^{n})^{T}\) by:
\begin{equation}
y_{1} = P(positive|x)
\end{equation}
\begin{equation}
y_{2} = P(negative|x)
\end{equation}

We use a one-layer CNN with the same hyperparameter settings as \citeauthor{kim2014convolutional} \shortcite{kim2014convolutional} but eliminate all the explicit regularization including \(L2\) norm and dropout. The three datasets we select are also used by \citeauthor{kim2014convolutional} \shortcite{kim2014convolutional} with pre-trained embedding. (1) \textbf{MR: } Movie reviews with one sentence per review which are either positive or negative \cite{pang2005seeing}. (2) \textbf{SST2: } Stanford Sentiment Treebank with neutral review removed and binary labels. \cite{socher2013recursive} (3) \textbf{Subj: } Subjectivity test for classifying a sentence is subjective or objective \cite{pang2004sentimental}. MR and Subj datasets do not provide a standard train/test split, thus 10-fold is used by strictly following the choice of \citeauthor{kim2014convolutional} \shortcite{kim2014convolutional} and \citeauthor{zhang2015sensitivity} \shortcite{zhang2015sensitivity}. SST2 has a train/test split, thus we average the performance over 10 runs.

\subsection{Language modeling}
Language modeling is the task of predicting the next word in a sentence. The model given current word \(w_{t}\) should learn to predict a probability distribution of the next word \(w_{t+1}\):
\begin{equation}
P(w_{t+1}|w_{t}; \theta)
\end{equation}
Using model to generate words is different from training a classifier in sentence classification. Thus, we choose it as our second task.

We use one dataset: Penn Treebank corpus \cite{marcus1993building} focusing only on a 2-layer LSTM. Learning rate decay is applied during training since it can significantly help the model converges better. To analyze the role of explicit regularization, focusing on dropout, we train the model in 2 different settings either without dropout or set it to 0.5. Both settings are repeated over 10 times. Consider about the effect of initialization, we choose 5 different initialization ranges: \([10^{-1}, 10^{-1.5}]\), \([10^{-1.5}, 10^{-2}]\), \([10^{-2}, 10^{-2.5}]\), \([10^{-2.5}, 10^{-3}]\), \([10^{-3}, 10^{-3.5}]\). Each of them has 2 fixed initialization schemes.

\subsection{Sequence tagging}
Sequence tagging, as a supervised learning task, is a core NLP problem. It can be recognized as a extension of classification problem and a simpler form of structure prediction. Input the word sequence and output the tagging sequence. Concretely, for an word sequence \(x = (x^{1}, x^{2},...,x^{n})^{T}\), we want to find the \(y = (y^{1}, y^{2},...,y^{n})^{T}\) with maximum conditional probability:
\begin{equation}
\max\limits_{y}P((y^{1}, y^{2},...,y^{n})^{T}|(x^{1}, x^{2},...,x^{n})^{T})
\end{equation}

We consider one dataset: CoNLL 2003 NER task \cite{tjong2003introduction}. The model architecture we use for training is BiLSTM with pretrained embedding and without conditional random field (CRF) loss since we want to consistently use ordinary cross entropy loss in all three tasks. We repeat the experiment five times and report the averaged results.

\section{Shallow Neural Network Experiments}
\begin{figure}
\centering
\includegraphics[width=7.5cm]{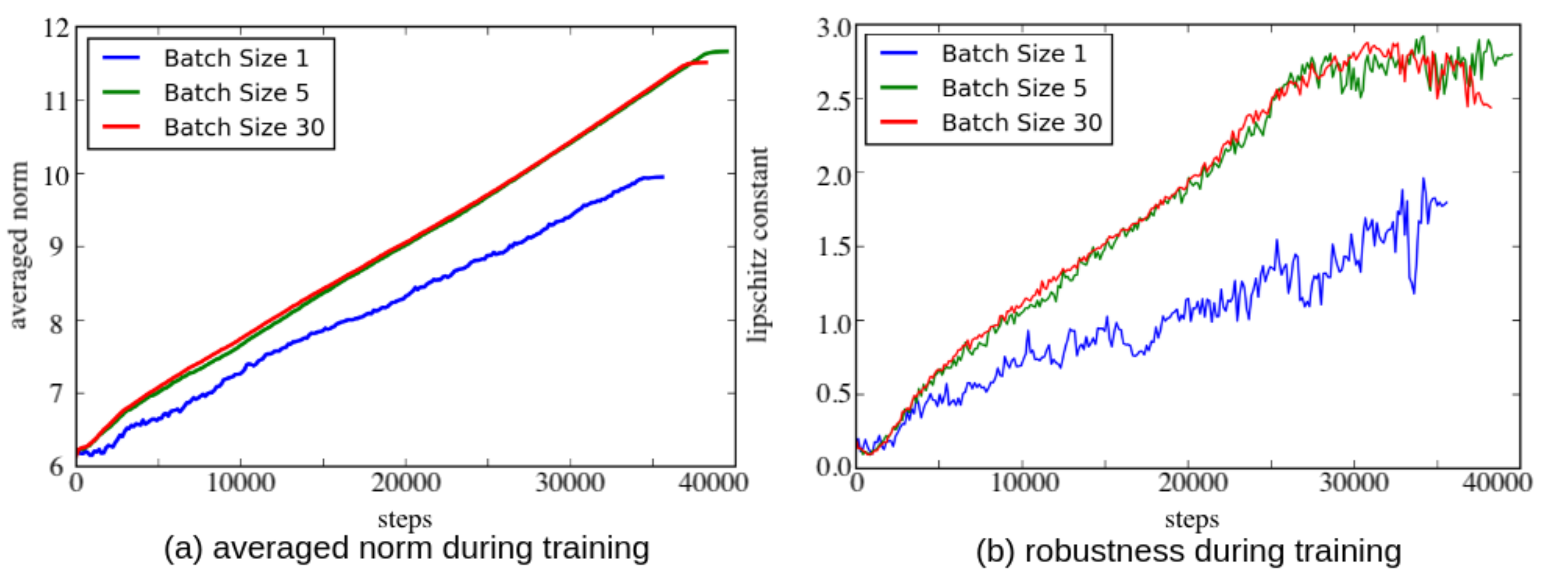}
\caption{(a) The averaged norm of all incoming/outgoing weight vectors. (b) Lipschitz constant for each model. Pure SGD outperforms others in both measures.}
\label{fig:norm}
\end{figure}

As deep neural networks remain mysterious for reasons, many researchers tried to reveal the inside logic starting from shallow models\cite{mianjy2018implicit,soudry2017implicit,gunasekar2017implicit}. It is useful to appeal to the simple case of shallow neural network models to see if there are parallel insights that can help us understand generalization better before we move on to the deep learning systems in the next section.

We build a simple 3-layer neural network on a non-linearly separable dataset as one step ahead from analyzing logistic regression on linearly separable data \cite{soudry2017implicit}. The dataset of size 200 with two interleaving half circles and 0.35 standard deviation of Gaussian noise is what we choose to let our model learn. The split ratio of train and test is 7:3.

The generalization capability of a model is evaluated by measuring the complexity of the model by specific yardsticks, where the \(L_{2}\) norm and the Lipschitz constant are two of the most widely used ones \cite{ng2004feature,xu2012robustness,mcallester1999pac}. Here, we choose these 2 methods to measure how implicit regularization is related to models' generalization as well:
\begin{itemize}
\item Averaged \(L_{2}\) norm of all weight vectors
\begin{equation}
\overline{L_{2}} = \frac{1}{n}\sum_{i=0}^{n}\sqrt{\sum_{j=0}^{p} \sum_{k=0}^{q} |W_{j,k}^{(i)}|^{2}}
\end{equation}
\item Lipschitz constant \(C\),  given loss function \(f: R \rightarrow R\) 
\begin{equation}
\exists C : \left |f(x) - f(y)\right | \leq C*L_{2} (x,y)
\end{equation}
\end{itemize}
Figure~\ref{fig:norm} shows pure SGD has the smallest averaged norm and Lipschitz constant due to the implicit regularization effect. Figure~\ref{fig:decision} shows the decision boundary of the model with different batch sizes. Along with the fact that all batch sizes achieves the same test accuracy, pure SGD with batch size 1 tends to have a more stable decision boundary that is prevented from over-fitting the data. Larger batch sizes are more volatile and prone to converge to a decision boundary with more curvature. Even though we select the simplest model with the best accuracy when measuring the performance on test data, the converging and over-fitting observation can give us some insight on how implicit regularization can help SGD achieve a better generalization.

The results increase our confidence that SGD's implicit regularization performance will be consistent on practical deep learning models. Our central findings in the 3-layer neural network can be summarized as:
\begin{itemize}
\item \textbf{F0-1:} SGD with smaller batch sizes have stronger implicit regularization effect as they approach closer to pure SGD.
\item \textbf{F0-2:} Pure SGD has far better implicit regularization performance, whereas any other batch sizes show similar performances.
\end{itemize}

\begin{figure}
\centering
\includegraphics[width=7.7cm]{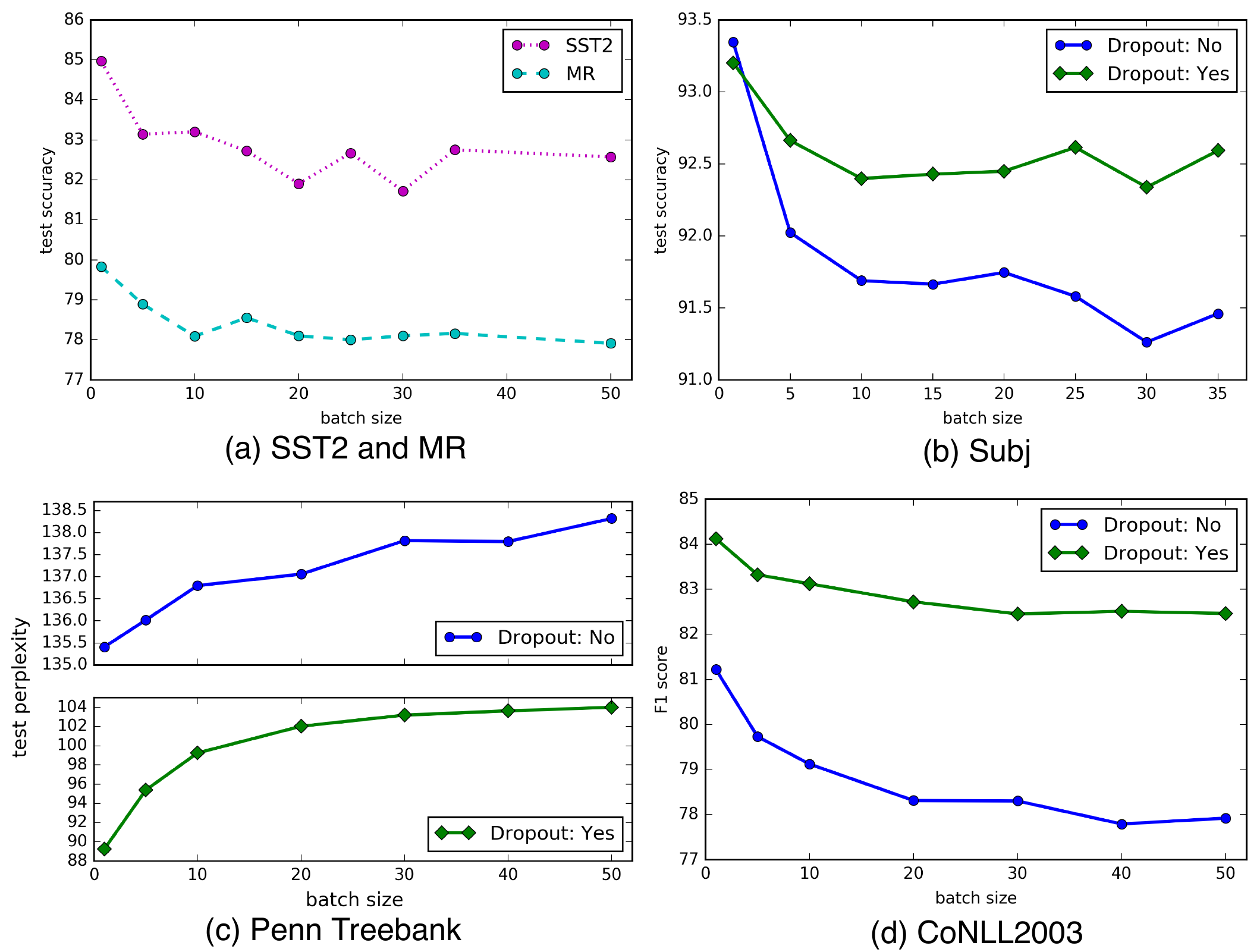}
\caption{The degrade of generalization performance with increased batch sizes is consistent across different tasks. Dropout cannot replace the SGD implicit regularization.}
\label{fig:tasksresult}
\end{figure}

\section{Analysis on Deep Learning}
We move on our study to deep learning systems in the following sections. To answer \textit{\textbf{RQ1}}, after eliminating all the explicit regularization effect, whether an over-parameterized model can have a good generalization performance is solely depends on the model architecture and its learning method. 

The results in Figure~\ref{fig:tasksresult} consistently reflect the findings in the shallow experiments. Pure SGD has a better generalization performance in all 3 tasks. Thus, we can generalize \textit{\textbf{F0-1}} and \textit{\textbf{F0-2}} to deep learning:

\smallskip
\noindent \textit{\textbf{F1-1:} generalization has a significant decrease between batch size 1 and 5 which shows even though the accuracy decreases almost monotonically, pure SGD still stands out among others.}
\smallskip

On the other hand, pure SGD converges significantly slower compared to others. Since current state-of-the-art models in NLP often use mini-batch methods to measure the performance, we recommend future studies use pure SGD to test models' generalization ability better, assuming researchers have appropriate computational resources.

\subsubsection*{Consistency of learning rate}
\begin{figure}
\centering
\includegraphics[width=7cm]{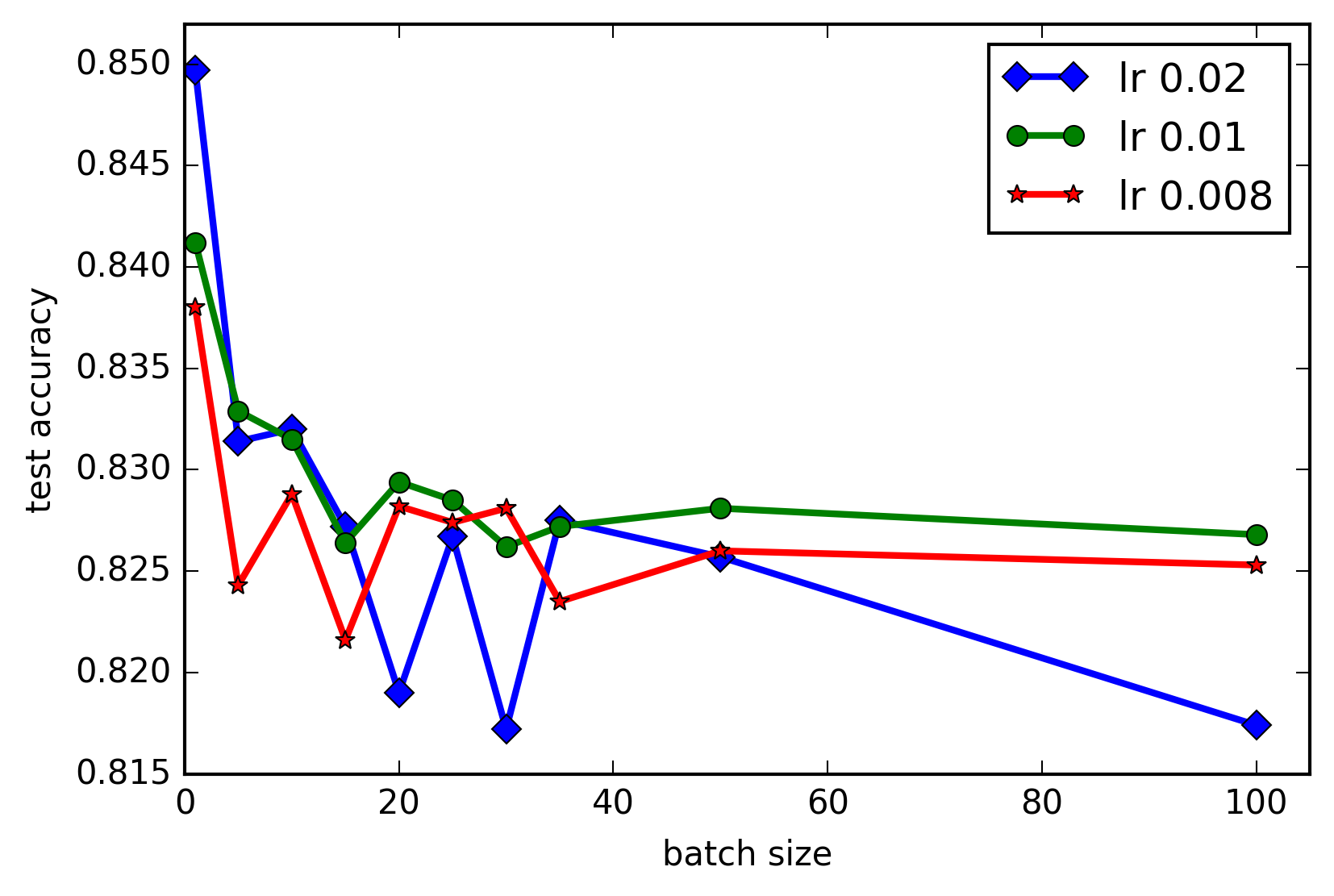}
\caption{'lr' in legend stands for learning rate. The degrade of performance while increasing batch size is consistent across different learning rates in SST2 dataset which makes the observation of implicit regularization solid.}
\label{fig:learningrate}
\end{figure}

Since using SGD, \citeauthor{kim2014convolutional} \shortcite{kim2014convolutional}'s CNN model can converge without learning rate decay. We try multiple learning rate to make sure the observation is convincing. The results of the SST2 dataset in Figure~\ref{fig:learningrate} show the effect is consistent. SGD with batch size 1 converges to a more generalizable global minima. Note that from Figure~\ref{fig:learningrate}, we can see the performance of batch size 1 is weakened as learning rate decreases because batch size 1 requires a relatively larger learning rate to converge better.

\subsection{The Role of explicit regularization}

\begin{table*}[t]
\scalebox{0.71}{
\centering
\begin{tabular}{|c|c|c|c|c|c|}
\hline
                                         &            & \multicolumn{2}{c|}{Dropout: No}                 & \multicolumn{2}{c|}{Dropout: Yes}                \\ \hline\hline
                                         & Batch Size & Train accuracy          & Test accuracy          & Train accuracy         & Test accuracy           \\ \cline{2-6}
\multirow{9}{*}{Sentence Classification} & 1          & 100.00                  & 93.35                  & 100.00                 & 93.20                   \\ 
                                         & 5          & 100.00                  & 92.02                  & 100.00                 & 92.66                   \\  
                                         & 10         & 100.00                  & 91.69                  & 100.00                 & 92.40                   \\  
                                         & 15         & 100.00                  & 91.66                  & 100.00                 & 92.43                   \\ 
                                         & 20         & 100.00                  & 91.75                  & 100.00                 & 92.45                   \\ 
                                         & 25         & 100.00                  & 91.58                  & 100.00                 & 92.61                   \\ 
                                         & 30         & 100.00                  & 91.26                  & 100.00                 & 92.34                   \\  
                                         & 35         & 100.00                  & 91.46                  & 100.00                 & 92.59                   \\ \hline\hline
                                         &            & Dev Perplexity          & Test Perplexity        & Dev Perplexity         & Test Perplexity         \\ \cline{2-6}
\multirow{7}{*}{Language Modeling}       & 1          & 140.08 (139.38,141.26)  & 135.41 (134.22,136.26) & 94.13 (93.83,94.30)    & 89.24 (88.80,89.59)     \\  
                                         & 5          & 139.73 (138.55,140.95)   & 136.02 (133.70,137.72) & 100.26 (99.90,100.87)  & 95.38 (94.76,95.96)     \\  
                                         & 10         & 140.47 (137.64,141.93)   & 136.80 (134.94,138.33) & 103.30 (102.24,103.92) & 99.25 (97.96,100.28)    \\  
                                         & 20         & 140.82 (138.85,142.38)   & 137.06 (135.38,138.14) & 105.58 (104.16,106.35) & 102.03 (100.66,102.65)  \\  
                                         & 30         & 141.60 (139.75,142.49)   & 137.82 (135.90,138.96) & 137.80 (134.89,139.55) & 138.23 (135.27,139.15)  \\ 
                                         & 40         & 141.45 (139.10,142.95)   & 137.80 (134.89,139.55) & 107.28 (106.18,107.95) & 103.63 (102.96,104.15)  \\ 
                                         & 50         & 141.29 (139.15,142.92)   & 138.32 (135.27,139.49)& 107.48 (106.11,108.10) & 104.00 (102.50,105.11)  \\ \hline\hline
                                         &            & Dev F1                  & Test F1                & Dev F1                 & Test F1                 \\ \cline{2-6}
\multirow{7}{*}{Sequence Tagging}        & 1          & 86.92 (80.83,81.60)     & 81.22 (80.83,81.60)    & 89.241 (88.89,89.74)   & 84.12 (83.53,84.50)     \\  
                                         & 5          & 85.72 (85.35,85.95)     & 79.73 (79.15,80.18)    & 88.598 (88.06,88.89)   & 83.32 (82.99,83.89)     \\ 
                                         & 10         & 85.09 (84.99,89.73)     & 79.12 (78.61,79.73)    & 88.401 (87.95,88.68)   & 83.12 (82.24,83.52)     \\  
                                         & 20         & 84.45 (84.19,84.76)     & 78.31 (77.50,78.68)    & 87.729 (87.56,87.86)   & 82.72 (82.21,83.23)     \\  
                                         & 30         & 84.20 (83.85,84.93)     & 78.30 (77.72,78.84)    & 87.594 (87.26,87.77)   & 82.45 (82.13,82.76)     \\  
                                         & 40         & 83.76 (83.47,84.16)     & 77.79 (77.28,77.99)    & 87.607 (87.54,87.78)   & 82.51 (82.14,82.74)     \\ 
                                         & 50         & 83.49 (83.29,83.80)     & 77.92 (77.53,78.43)    & 87.619 (87.50,87.93)   & 82.46 (82.20,82.94)     \\ \hline
\end{tabular}}
\caption{SGD implicit regularization in 3 tasks. It exists both with and without dropout which leads to finding 3. We use Subj in sentence classification, Penn Treebank in language modeling and CoNLL2003 in sequence tagging. Since Subj does not have a development set, 10-fold cross-validation is used. We select dropout rate of 0.2 for subj and 0.5 for the rest two. The implicit regularization exists both with and without dropout in all three tasks.}
\label{tab:taskresults}
\end{table*}

Most of our experiments are done without explicit regularizations. These strategies are used to help deep neural nets with a large number of parameters deal with overfitting. We specifically focus on dropout in this paper, since it has a greater impact on other regularization methods \cite{srivastava2014dropout}. The key idea is to randomly drop units during training to prevent units co-adapting too much.

We do have an understanding of the implicit bias of dropout that tends to make the norm of weight vectors of all hidden units equal for single hidden-layer linear neural networks \cite{mianjy2018implicit} and its close connection with \(L_{2}\) regularizer \cite{wager2013dropout,helmbold2015inductive}. Empirical study shows that even though dropout and \(L_{1}\), \(L_{2}\) regularization are important, model architecture modification is more crucial \cite{zhang2015sensitivity,zhang2016understanding}.

Our experiments confirm that dropout could potentially improve generalization performance. However, more importantly, sentence classification, language modeling and sequence tagging experiments with dropout can still observe the implicit regularization. We further compare the performance of implicit regularization with and without dropout in all three tasks. The result showed in Figure~\ref{fig:tasksresult} implies that the implicit regularization effect is weakened but still exists which keeps pure SGD as the most generalizable model.

\smallskip
\noindent \textit{\textbf{F1-2:} Dropout, as an explicit regularization, neither tremendously affect the generalization performance nor completely substitute the regularization effect of SGD.}
\smallskip

We recommend researchers further study the impact of other explicit regularization such as early stop and L2 regularizer. Although dropout is only a small piece of the explicit generalization story, our study points out a direction that implicit regularization is the major reason why over-parameterized model albeit can learn all the training data by brutal force, but still generalizes well.

\section{Neural Network's Finite Capability}
Deep learning model has its own finite capability of understanding state-of-the-art datasets. Under rare circumstances, the intrinsic pattern of the training data is hard to discover, which leads to a challenge for whether over-parameterized model will learn the data by brute-force or reach its limit to understanding the pattern. We study how implicit regularization is affected by the dataset challenges. Specifically, we test our models on both training data that is smaller than the original and with corrupted labels which leads to a condition that is almost impossible to learn. We answer our \textit{\textbf{RQ2}} here.

\subsection{Limited sample expressivity}
\begin{figure}[ht]
\centering
\includegraphics[width=7.7cm]{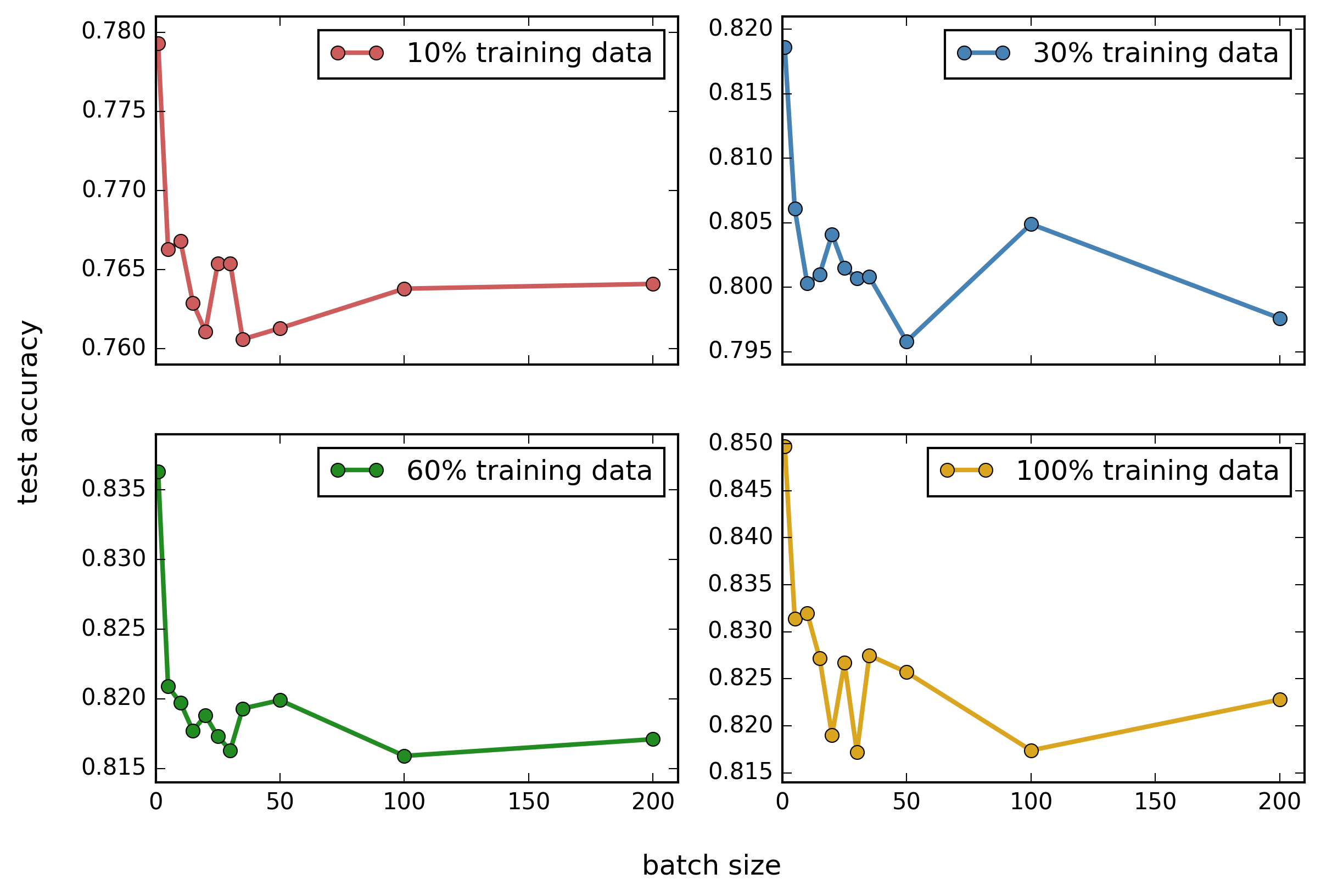}
\caption{The degrade of generalization performances with increased batch size trained by subsets of the original SST2 training data. The selected subset size of the original training sample only have a small impact on the performance degrade while increasing batch size.}
\label{fig:datasize}
\end{figure}
There exists some analysis for the expressivity of neural networks \cite{mhaskar2016deep,cohen2016convolutional}. One recent study on finite sample size is: how large the model should be in order to learn a certain number of random labels by brutal force \cite{zhang2016understanding}. Here, we further generate a contrary question about the analysis: Is there a specific limitation about the size of data so that neural networks can hardly capture the real pattern and have difficulty choosing whether to fit as random labels easily or try their best to find intrinsic features? Admittedly, it is more realistic to use SGD if the training sample size is small, as SGD is the slowest optimization algorithm comparing with Adadelta \cite{zeiler2012adadelta}, Adagrad \cite{duchi2011adaptive}, and Adam \cite{kingma2014adam}. However, if this limitation is not small enough, we cannot utilize the implicit regularization ability of SGD on models with small data size. Verifying the how models' expressivity affect SGD implicit regularization on limited sample dataset is crucial.

In the sentence classification experiment on the SST2 dataset, we extract 0.1, 0.3, 0.6 percent of the original data. Intuition suggests that if the dataset size is extremely small, the neural networks will struggle to learn the intrinsic pattern from the data, namely, extremely bad generalization performance. Unexpectedly, even though the performance of the model steadily decreases as the decreasing of the data percentage, we can observe that the implicit regularization effect is consistent under different percentages of the original training data. An important conclusion can be made:

\smallskip
\noindent \textit{\textbf{F2-1:} implicit regularization effect is one of the intrinsic properties of SGD which will not be affected by data size for over-parameterized models.}

\begin{figure}[t!]
\centering
\includegraphics[width=7cm]{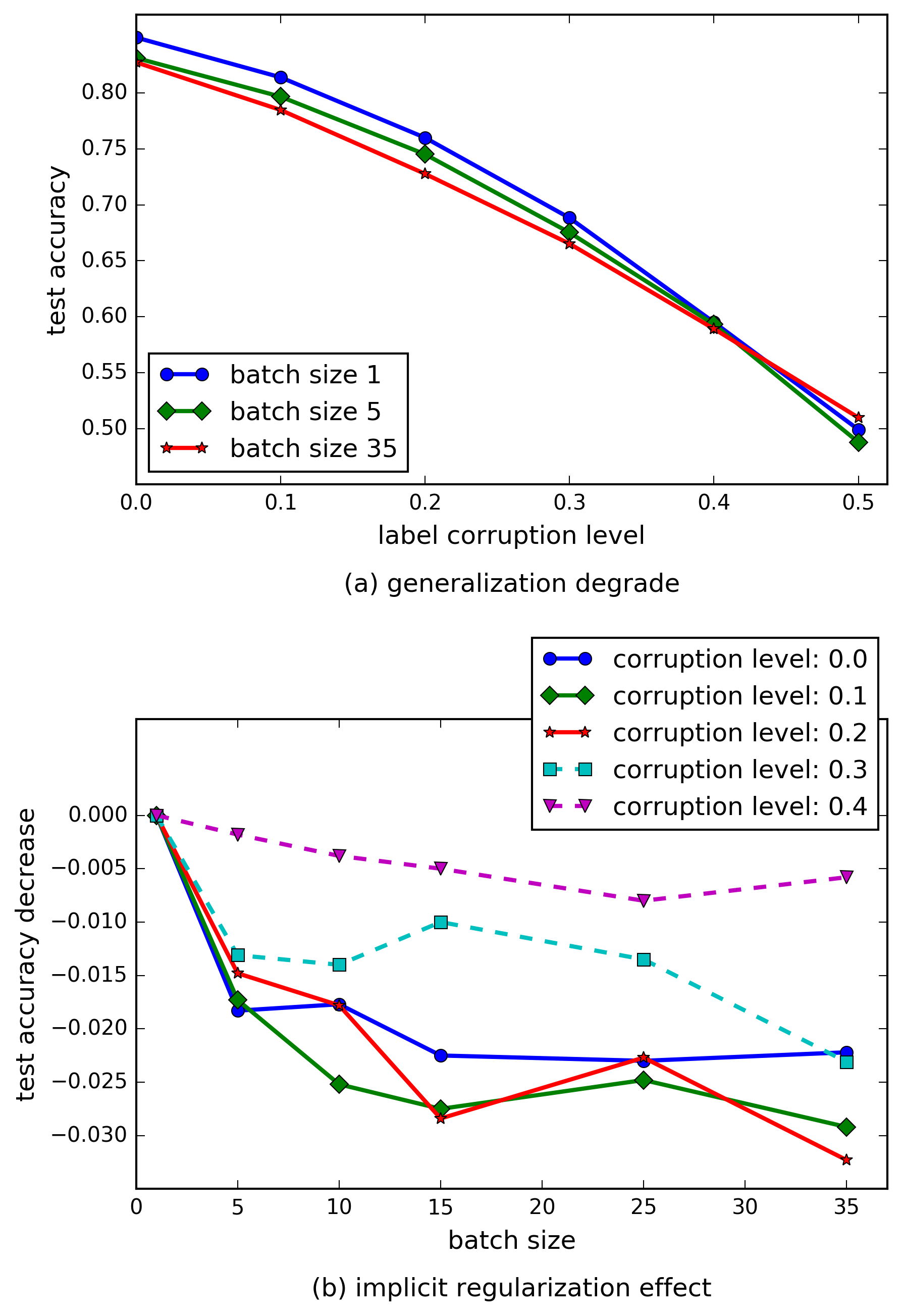}
\caption{Fitting corrupted labels on SST2. (a) shows relative performance degrade with different corruption levels. Generalization gap for different batch sizes is maintained before corruption level 3. (b) shows the gap of accuracy comparing to pure SGD as the baseline. implicit regularization effect is similar for corruption level smaller than 3.}
\label{fig:corruption}
\end{figure}

\begin{table*}[t]
\scalebox{0.77}{
\centering
\begin{tabular}{c||ccccccc}
 	\hline
    Initialization range & Batch size 1 	& Batch size 5 			& Batch size 10 		& Batch size 20 		& Batch size 30 		& Batch size 40 		& Batch size 50 \\ \hline
    \([10^{-1.5}, 10^{-1}]\)                & 135.11      			& 135.46      			& \fbox{134.94}         & \fbox{135.28}       & \fbox{135.90}       & \textbf{\fbox{134.89}}       & \fbox{135.27}\\ \hline
    \([10^{-2}, 10^{-1.5}]\)                & 136.26        		& \textbf{\fbox{133.70}}    	& 136.46         		& 135.83        		& 138.59        	    & 136.94                & 138.83         \\ \hline
    \([10^{-2.5}, 10^{-2}]\)                & \textbf{\fbox{134.22}}       & 136.03                & 138.33         		& 138.14        		& 137.37        		& 139.10                & 139.13         \\ \hline
    \([10^{-3}, 10^{-2.5}]\)                & \textbf{136.23}		        & 137.72        		& 137.00        		& 137.93        		& 138.32        		& 138.51        		& 138.77         \\ \hline
    \([10^{-3.5}, 10^{-3}]\)                & \textbf{135.24}	     	    & 137.22        		& 137.29        		& 138.02        		& 138.96        		& 139.55        		& 139.15         \\ \hline
\end{tabular}}
\caption{The effect of initialization range in Penn Treebank dataset. Implicit regularization is weakened for larger initialization ranges. Smaller batch size prefers small initialization setting.}
\label{tab:initrange}
\end{table*}

\subsection{Fitting corrupted labels}
We run our experiments with partially corrupted labels as a variant of the randomization test from non-parametric statistics \cite{edgington}. For a corrupted level \(p\), Each training label has a probability \(p\) to be corrupted as a wrong label independently. The sentence classification datasets we choose have binary labels, thus the corrupted labels will be flipped to the wrong ones. This study is inspired by \citeauthor{zhang2016understanding}'s generalization study \shortcite{zhang2016understanding}. He surprisingly observed SGD on several state-of-the-art models could fit any level of label corruption including random labels that challenges the traditional thinking of generalization. Even though the training error can be reduced to 0, the generalization ability is still affected by the label corruption due to over-fitting.

Since the model tends to recognize the intrinsic pattern of the training sample while at the same time brutal force to learn the rest, why over-parameterized model still has generalization ability while it could brutal force to learn all the data is unknown. One interesting observation is that the performance of the model with larger batch sizes, such as 35 in our sentence classification experiments, does not decrease suddenly at some corruption levels but decrease predictably based on the percentage of corruption. On the contrary, SGD with smaller batch sizes tend to keep the generalization gap with large batch sizes and start to lose this advantage at some label corruption levels close to 0.3. All batch sizes converge to almost same generalization performance at corruption level close to 0.4. Since corruption level 0.5 on a binary label dataset means the model will do random prediction, which makes different batch sizes the same. However, the question is why this kind of convergence cross-batch sizes only happens after some corruption level? Why smaller batch sizes have a larger decreasing rate?

We believe there are some intrinsic reasons that prone small batch size models especially pure SGD to learn the pattern as much as possible and the implicit regularization is one of the biggest. To gain further insight into this phenomenon, we measure the role of implicit regularization on multiple corruption levels. It can be clearly seen that the implicit regularization effect remains strong at corruption level 0.1 and 0.2, but weakened at 0.3.

\smallskip
\noindent \textit{\textbf{F2-2:} SGD implicit regularization improves generalization while brutal force learning is required.}

It gives us new insights into over-parameterized models' generalization ability. Although implicit regularization of SGD might not completely solve the intriguing question of generalization, it sheds lights on future study of deep learning model's intrinsic ability and the strategy of optimization.

\section{Weight Initialization Stability}
The widespread use of various kinds of initialization steps inside the mainstream of deep learning studies \cite{saxe2013exact,sutskever2013importance,duchi2011adaptive}. The Xavier Initialization and the He-et-al Initialization \cite{jia,he2015delving} are the widely used. The study of \citeauthor{sutskever2013importance} \shortcite{sutskever2013importance} shows that well-designed initialization can bring a breakthrough to the performance of deep and recurrent neural networks using SGD. Consider the role of initialization in deep learning, verifying the consistency of our observation, pure SGD has strongest implicit regularization among other batch sizes, is necessary. We present our answer for \textit{\textbf{RQ3}} based on Table~\ref{tab:initrange}. In language modeling, a smaller initialization range, there exists a strong effect but large initialization of the weights can weaken it. This observation is consistent with a recent theory study on matrix factorization \cite{li2017algorithmic} that 

\smallskip
\noindent \textit{\textbf{F3:} Smaller initialization range will lead to a better generalization effect.}
\smallskip

In addition, a large batch size might prefer a larger initialization range albeit it can not guarantee a globally best solution. Based on the above study, using a slightly larger batch size such as 5 with a proper weight initialization setting could potentially restore the implicit regularization while converges a lot faster. We will continue the study of how to make the effect more practical.


\section{Related work}
A few papers have constructed theories of SGD implicit regularization. \citeauthor{li2017algorithmic} \shortcite{li2017algorithmic} prove the implicit regularization effect of SGD on small machine learning models based on the fact that SGD prefers some generalizable local minimas to others \cite{srebro2011universality}. However, these theories does not establish universal conclusions which can be applied to deeper neural networks with more practical activations and the cross-entropy loss function. Such theory construction requires a better understanding to optimization algorithm's behavior for non-linear models and statistics. In this work, we study the unsolved problem base on large-scale and comprehensive empirical studies.

One study shows that the logistic regression model and 2-layer neural networks using monotone decreasing loss functions tend to converge in the direction of the max-margin solution when using GD and SGD \cite{soudry2017implicit}. We further enhance the conclusion by doing studies on more practical deep learning systems.

\citeauthor{zhang2016understanding} \shortcite{zhang2016understanding} point out how traditional approaches fail to describe why over-parameterized models generalize well by successfully letting state-of-the-art models fit corrupted labels. However, this study does not have a clear conclusion for the implicit regularization effect. In this work, we explore further on how implicit regularization is affected by over-parameterized model's capability of fitting corrupted labels. The role of dropout in SGD's implicit regularization is also analyzed based on the work of \citeauthor{wager2013dropout} \shortcite{zhang2016understanding} and \citeauthor{mianjy2018implicit} \shortcite{mianjy2018implicit}.

\section{Conclusions}
We analyze the effect of SGD implicit regularization in the deep learning systems, especially in NLP tasks. Results are consistent across all kinds of settings: (i) Dropout, as an explicit regularizer, cannot completely replace the effect of SGD implicit regularization. (ii) Models' limited capabilities do not greatly affect the SGD implicit regularization.  (iii) SGD models with stronger implicit regularization tend to prefer lower initialization ranges. Our study can be utilized to help future researchers to measure their state-of-the-art models' generalization performance better by using SGD.

\bibliography{paper} 
\bibliographystyle{paper}

\end{document}